\documentclass[a4paper]{svproc}

\pdfoutput=1
\usepackage{url}

\usepackage{graphicx}
\usepackage{siunitx}
\usepackage{subcaption}
\usepackage{hyperref}
\usepackage{amsmath}
\usepackage{amsfonts}
\usepackage{amssymb} %
\usepackage{cite}
\usepackage{bm}
\usepackage{xcolor}
\usepackage[ampersand]{easylist}
\usepackage{wrapfig}
\usepackage[font=small,labelfont=bf]{caption}
\usepackage{cleveref} %
\usepackage{bm}
\usepackage{mathtools} %

\usepackage{algorithm}
\usepackage{algpseudocode}

\newcommand{\fa}{\bm{a}}
\newcommand{\fb}{\bm{b}}
\newcommand{\fA}{\mathcal{A}}
\newcommand{\fB}{\mathcal{B}}
\newcommand{\fI}{\mathcal{I}}
\newcommand{\fJ}{\mathcal{J}}
\newcommand{\fL}{\mathcal{L}}

\newcommand{\Tal}{\mathbf{T}_{\mathcal{A}\mathcal{L}}}
\newcommand{\Tbl}{\mathbf{T}_{\mathcal{B}\mathcal{L}}}

\newcommand{\qAB}{\mathbf{q}_{\fA\fB}}

\newcommand{\qBb}{\mathbf{q}_{\fB\fb}}
\newcommand{\qAa}{\mathbf{q}_{\fA\fa}}

\let\amssymbboxplus\boxplus
\let\amssymbboxminus\boxminus

\renewcommand{\boxplus}{\mathbin{\mathop\amssymbboxplus}}
\renewcommand{\boxminus}{\mathbin{\mathop\amssymbboxminus}}

\crefname{figure}{Fig.}{Figs.}
\crefname{section}{Section}{Sections}

\usepackage[nolist,nohyperlinks]{acronym}
\begin{acronym}
\acro{PCC}{Pearson Correlation Coefficient}
\acro{MAE}{Mean-Absolute-Error}
\acro{KL}{Kulback-Leibler}
\acro{CM}{Cram\'er von Mises}
\acro{KDE}{Kernel Density estimate}
\acro{IMU}{Inertial Measurement Unit}
\acro{RTK}{Real Time Kinematics}
\acro{NDT}{Non-Destructive Testing}
\end{acronym}

\begin{document}

\title{To Fuse or Not to Fuse: Measuring Consistency in Multi-Sensor Fusion for Aerial Robots}
\titlerunning{To Fuse or Not to Fuse?}
\author{Christian Lanegger \and Helen Oleynikova \and Michael Pantic \and Lionel Ott \and  Roland Siegwart}
\authorrunning{Christian Lanegger et al.} %
\tocauthor{Christian Lanegger, Helen Oleynikova, Michael Pantic, Lionel Ott, Roland Siegwart}
\institute{Autonomous Systems Lab, ETH Z\"{u}rich, Switzerland\thanks{This work was supported by the Hilti Group and by NCCR Digital Fabrication.}\vspace{-20pt}
}
\maketitle              %

\begin{abstract}
Aerial vehicles are no longer limited to flying in open space: recent work has focused on aerial manipulation and up-close inspection. Such applications place stringent requirements on state estimation: the robot must combine state information from many sources, including onboard odometry and global positioning sensors. However, flying close to or in contact with structures is a degenerate case for many sensing modalities, and the robot's state estimation framework must intelligently choose which sensors are currently trustworthy.

We evaluate a number of metrics to judge the reliability of sensing modalities in a multi-sensor fusion framework, then introduce a consensus-finding scheme that uses this metric to choose which sensors to fuse or not to fuse.
Finally, we show that such a fusion framework is more robust and accurate than fusing all sensors all the time and demonstrate how such metrics can be informative in real-world experiments in indoor-outdoor flight and bridge inspection.
\end{abstract}

\section{Introduction}
\vspace{-5pt}
The capabilities of aerial robots have continuously expanded over the last years thanks to improved hardware and increased computing power. As a result, new applications such as non-destructive contact-based inspection \cite{2021aerialmanip}, window cleaning \cite{window2021sun}, or precise writing \cite{Tzoumanikas2020writing} have come within reach for aerial systems. 
However, operating close to structures as opposed to in free space creates many new challenges for aerial robots. State estimation has to be significantly more precise, as an error of a few centimeters is the difference between being next to a wall and crashing into it. 
In such applications, combining multiple sources of information and thus coordinate frames, such as absolute GPS position, relative position to a structure, and other external surveying stations, is often necessary.
Many of these sensors are also subject to degraded performance or dropouts due to occlusions.
The traditional intuition that more sensors are always better and fusing as much data as possible from as many sources as possible fails to hold in these cases, and estimating which sensors are currently reliable becomes indispensable.
This work introduces a metric that can measure the consistency of multiple odometry, pose, and position sensor estimates. Based on this metric, we design an algorithm that decides which sensors to fuse at any point in time.

To leverage the ability to selectively fuse sensors based on their consistency, we propose a parallel-redundant state estimation framework that treats each sensor as an individual process. For each sensor, process metrics that can detect noise, dropouts, or other sources of inconsistencies are defined, enabling the system to estimate which sensors are currently reliable and fuse only those into the final estimate.
This results in a robust and accurate state estimator even in challenging scenarios such as flight near surfaces. In summary, our contributions are as follows:
\begin{easylist}[itemize]
    & Evaluation of multiple metrics for multi-sensor consistency estimation.
    & Introduction of an effective sensor fusion framework in the presence of dropouts and divergence.
    & Real-world experimental evaluation in challenging scenarios.
\end{easylist}

\vspace{-5pt}
\subsection*{Related Work}
\vspace{-5pt}
Most state-of-the-art state estimation approaches use a unified filtering or smoothing framework to estimate the full global state of the system \cite{2020fullyactuatedcontactinspection}. For a minimal sensor setup tailored to a specific environment, such an approach works well. With increasingly complex systems, the number of correlated states that need to be co-estimated, some of which are not directly observable, increases and can result in the optimization getting stuck in local minima. Filtering methods can become overconfident or diverge in the presence of outliers or irregular sensor readings. Recently, \cite{mars2021weiss} tried to address this issue by covariance segmentation, allowing filter propagation and updates on a per-sensor basis. Nubert \textit{et al.} 
 addressed the issue of GNSS dropouts by switching between two optimization problems~\cite{nubert2022graph}. 
Others try to increase resilience to degenerate sensing by directly looking at raw sensor input (such as images) \cite{Brunner2014} or degenerate environments (such as LiDAR scans) \cite{nubert2022localizability} and define metrics that determine if a measurement is used. 
\cite{lee2023mins} takes the opposite approach and fuses all available data; however, their results show that there are cases where this leads to decreased accuracy.
In contrast to these approaches, we propose a metric that's sensor-invariant, as it acts on pose or position estimates rather than raw data. We also only \textit{selectively} fuse sensors that we have a high confidence in, based on our metric.

\vspace{-5pt}
\section{Technical Approach}
\vspace{-5pt}

This section introduces the three components of our approach: the \textit{metric} measuring the consistency between estimates from different sensors, the \textit{state estimation framework} fusing multiple sensors together, and finally, the approach using the metric in the state estimator to choose which sensors to fuse by finding \textit{consensus} among them.

\vspace{-8pt}
\subsection{Consistency Metric}
\vspace{-5pt}
The main goal of this work is to find an efficient consistency metric that quantifies the mutual information between measurements of at least two sensors. The metric should detect inconsistencies between measurements of multiple sensors, allowing us to detect sensor failures or spurious state estimates effectively. To determine if two sensors, rigidly attached to the robot at $\bm{a}$ and $\bm{b}$ with local reference frames $\mathcal{A}$ and $\mathcal{B}$ respectively, provide similar information about the robot's state, we first estimate the homogeneous transformations between the sensors' local reference frames $\Tal$ and $\Tbl$. Then, we compute the linear velocity at $\bm{a}$ with respect to $\mathcal{A}$ as measured by sensor $\bm{b}$:
\begin{equation}
    {}_{\fa}\mathbf{v}_{\fa}^{\fb} = \qAa^{-1}\circ \qAB \circ \qBb \left({}_{\fb}\tilde{\mathbf{v}}_{\fb} + \boldsymbol{\tilde{\omega}}_{\fb} \times {}_{\fb}\mathbf{r}_{\fb\fa}
    \right),
\end{equation}
\begin{wrapfigure}{r}{0.4\textwidth}
    \vspace{-20pt}  
    \centering
    \includegraphics[width=1.0\linewidth]{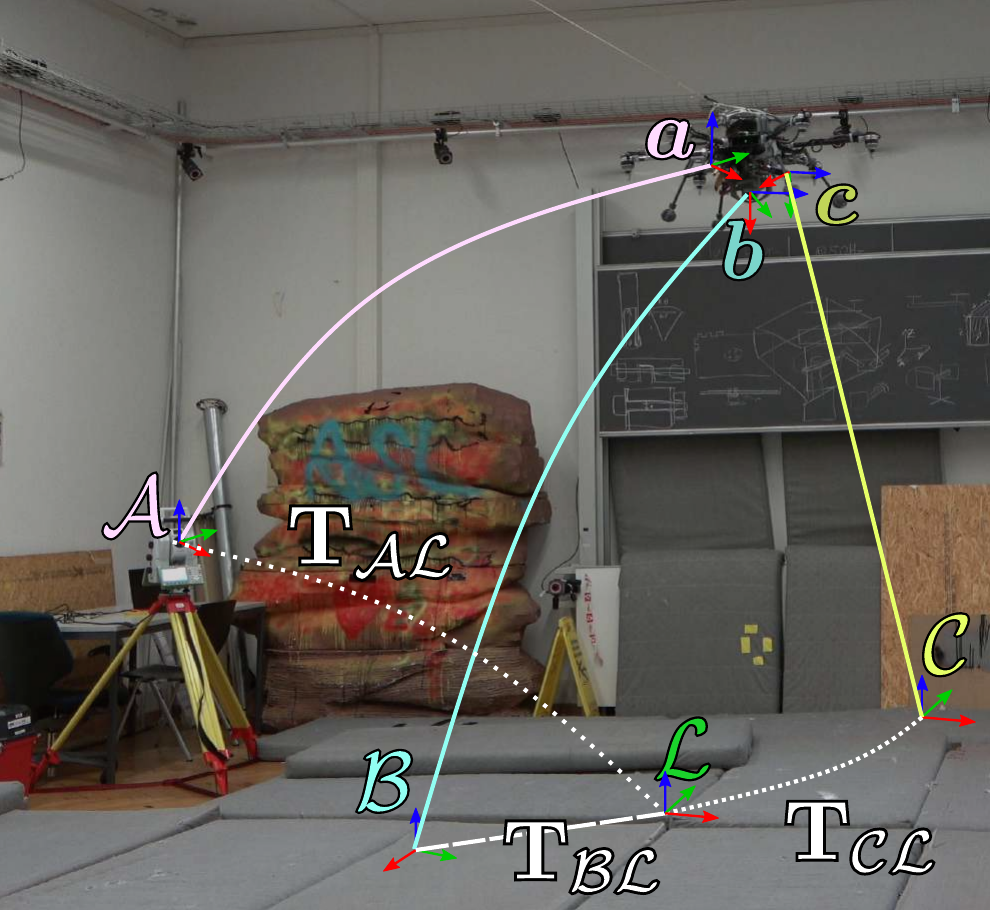}
    \caption{Coordinate frames of multiple sensors, $\bm{a},\bm{b},\bm{c}$ relative to their global frames $\mathcal{A},\mathcal{B},\mathcal{C}$, where we estimate the transformation between these frames and the local inertial frame $\mathcal{L}$.}
    \vspace{-20pt}
    \label{fig:reward}
\end{wrapfigure}
where $\qAa \in SO\left(3\right)$ denotes the rotation from $\fA$ to $\fa$,  ${}_{\fb}\tilde{\mathbf{v}}_{\fb}$ is the velocity as measured by sensor $\fb$, $\boldsymbol{\tilde{\omega}}_{\fb}$ the robot's angular velocity w.r.t. $\fb$, ${}_{\fb}\mathbf{r}_{\fb\fa}$ the intrinsic calibration between $\fb$ and $\fa$, and $\circ$ the concatenation operation. Each component of the estimated velocity vector ${}_{\fa}\mathbf{v}_{\fa}^{\fb}$ is then compared to the component of ${}_{\fa}\mathbf{v}_{\fa}^{\fa}$, the measured linear velocity of sensor $\fa$,  using a distance metric. The distance metrics for each axis are summed together to obtain a final consistency value.
We evaluate the following potential metrics:

\paragraph{\ac{MAE}:} The \ac{MAE} is defined as the mean absolute error between the $n$ most recent velocity components ${}_{\fa}v_{\fa}^{\fa}$ and ${}_{\fa}v_{\fa}^{\fb}$:
\small
\begin{equation}
D_{MAE} = \frac{\sum_{i=1}^n \left\| {}_{\fa}^{i}v_{\fa}^{\fa} - {}_{\fa}^{i}v_{\fa}^{\fb} \right\|}{n}
\end{equation}
\normalsize
\paragraph{Pearson Correlation Coefficient (PCC):} The PCC measures the linear relationship between two samples ${}_{\fa}v_{\fa}^{\fa}$ and ${}_{\fa}v_{\fa}^{\fb}$ with mean ${}_{\fa}\bar{v}_{\fa}^{\fa}$ and ${}_{\fa}\bar{v}_{\fa}^{\fb}$, respectively. It is defined as the ratio between the covariance of these two random variables and the product of their respective standard deviations:
\small
\begin{equation}
D_{PCC} = \frac{\sum ^n _{i=1}(x_i - \bar{x})(y_i - \bar{y})}{\sqrt{\sum ^n _{i=1}(x_i - \bar{x})^2} \sqrt{\sum ^n _{i=1}(y_i - \bar{y})^2}}
\end{equation}
\normalsize
\paragraph{Kulback-Leibler Divergence (KL):} The KL measures the amount of mutual information between two distributions.
For two discrete probability distributions $P$ and $Q$ the KL-Divergence is defined as:
\small
\begin{equation}
    D_{KL} = \sum_{x \in \mathcal{X}}P(x)\log{\frac{P(x)}{Q(x)}}\;.
\end{equation}
\normalsize
\paragraph{\ac{CM}:} The \ac{CM}-Distance is defined between two empirical cumulative distribution functions $F_P(x)$ and $F_Q(x)$, corresponding to their respective distributions $P$ and $Q$, as follows:
\small
\begin{equation}
D_{CM} = \left(2\int_{-\infty}^{\infty} \left(F_P(x) - F_Q(x)\right)^2\right)^{1/2}\;.
\end{equation}
\normalsize
This metric can be applied to measure the distance between a sample and a distribution or between multiple samples. In contrast to the KL-Divergence, the \ac{CM}-Distance considers the metric space between two distributions and is scale-sensitive.
For a more thorough comparison of the KL-Divergence and the \ac{CM}-Distance, the reader is referred to \cite{bellemare2017cramer}. 

Using body velocity rather than position for metric estimation decouples the effect of different reference frames, as \textit{body velocity} is a natural common frame for all estimators.
Using the differentiated position, we also avoid the effects of potentially integrating noise and have a faster response to changes.

\vspace{-5pt}
\subsection{Loosely coupled multi-graph optimization}
\vspace{-5pt}
Our state estimator consists of multiple loosely coupled factor graphs implemented using the GTSAM framework \cite{gtsam}, as used previously in \cite{laneggerpantic2023chasing}. A first graph estimates the robot's local state using only states constrained by relative measurements coming from an \ac{IMU} and one additional sensor providing pose or position measurements. The relative nature of the constraints allows seamless switching between measurement sources at the cost of accumulating errors over time. To account for the resulting drift, an additional factor graph is added for each sensor that estimates the transformation $\mathbf{T}_{\fA\fL}$ between a sensor's reference frame $\mathcal{A}$ and the robot's local reference $\mathcal{L}$, as visualized in \cref{fig:reward}. To estimate this transformation $\mathbf{T}_{\fA\fL}$, each sensor graph uses the local estimate of the local graph and the measurement of the sensor. 

\subsubsection{Local State Estimation.}
The state of the local graph at a given time is
\begin{equation}
{}_{i}\mathbf{x} \coloneqq
\left[
\mathbf{q}_{\fL\fI} , {}_{\fL}\mathbf{p}_{\fL\fI} , {}_{\fI}\mathbf{v}_{\fI}, {}_{\fI}\mathbf{b}_{a}, {}_{\fI}\mathbf{b}_{g}
\right]
\end{equation}
where ${}_{\fL}\mathbf{p}_{\fL\fI} \in \mathbb{R}^3$ represents the position of the \ac{IMU} frame $\fI$ relative to $\fL$. A new state ${}_{\fI}\mathbf{x}$ is added to the graph with every \ac{IMU} measurement, constraining the previous state to the new one. An \ac{IMU} factor is added to the optimization cost as an additive term. For a detailed definition, the reader is referred to \cite{forster2016manifold}. Pose or position measurements are added as \textit{between factors}, representing the error between the predicted and measured relative displacement in pose or position, respectively. By defining the body-fixed frame at state ${}_{i}{\mathbf{x}}$ as $\mathcal{I}$ and the same frame at a later state ${}_{j}{\mathbf{x}}$ as $\mathcal{J}$ the additive error term can be formulated as:
\begin{equation}
    r_{\fI\fJ}^o = \omega_o \left\| 
    {}_{\fL}\mathbf{p}_{\fL\fJ} - {}_{\fL}\mathbf{p}_{\fL\fI} - \mathbf{q}_{\fL\fI}\left( {}_{\fI}\bar{\mathbf{p}}_{\fI\fJ} \right) \right\|_2^2 
    + 
    \frac{\omega_q}{2} \left\| \mathbf{q}_{\fL\fJ} \boxminus \left( \mathbf{q}_{\fL\fI} \circ \bar{\mathbf{q}}_{\fI\fJ} \right) \right\|^2_F,
\label{eq:pose_between_residual}
\end{equation}
where $\left\|\cdot\right\|_F$ is the Frobenius norm, $\left\|\cdot\right\|_2$ is the $L_2$-norm, and $\omega_p$ and $\omega_q$ are concentration parameters. Sensor measurements are denoted with a bar on top. Similarly, for two position measurements $\fa$ and $\fb$ with respect to the reference frame $\fA$, the residual for the positional displacement is given by:
\begin{equation}
    r_{\fI\fJ}^p = \omega_p \left\| 
     {}_{\fL}\mathbf{p}_{\fL\fI} + \mathbf{q}_{\fL\fI} \left( {}_{\fI}\mathbf{p}_{\fI\fa}\right) -  
     {}_{\fL}\mathbf{p}_{\fL\fJ} + \mathbf{q}_{\fL\fJ}\left( {}_{\fJ}\vec{p}_{\fJ\fb}\right) -
    \mathbf{q}_{\fA\fL}^{-1}\left( {}_{\fA}\bar{\mathbf{p}}_{\fa\fb}\right)
    \right\|^2_2,
\end{equation}
where $\mathbf{q}_{\fA\fL}$ is the sensor transformation estimated by the graph optimization. The output of the local graph is used directly by the platform's controller and, therefore, has to be locally consistent and available at a high frequency. Trading accuracy for speed, we optimize the local graph at \SI{30}{\hertz} and use a fixed-lag smoother with a relatively small window size of \SI{0.5}{s}. Between optimizations, the state is integrated using \ac{IMU} measurements and output at \SI{200}{\hertz}.

\subsubsection{Sensor Transform Estimation.}

A separate optimization problem is set up for each sensor to estimate the drift between the sensor and local estimate, defined as a relative transformation between their corresponding references $\fA$ and $\fL$. The state is given as:
\begin{equation}
{}_{\fa}\boldsymbol{x} \coloneqq
\left[
\mathbf{q}_{\fA\fL} , {}_{\fA}\mathbf{p}_{\fA\fL}
\right],
\end{equation}
and constrained using the following position residual
\begin{equation}
r_{\fA\fL}^t = \omega_t
     \left\| 
 {}_{\fA}\bar{\mathbf{p}}_{\fa} - 
 \left( {}_{\fA}\mathbf{p}_{\fA\fL} + \mathbf{q}_{\fA\fL}\left({}_{\fL}\mathbf{p}_{\fa} \right)\right)
 \right\|_{2}^2,
\end{equation}
where ${}_{\fA}\bar{\mathbf{p}}_{\fa}$ is the measurement provided by the sensor and $\mathbf{q}_{\fA\fL}$ is obtained from the estimate of the local graph. Whenever position measurements are used as constraints locally, the estimated states of the local graph and the sensor transform graph become co-dependent. Therefore, in such a case, the sensor transform optimization is stopped and $\mathbf{T}_{\fA\fL}$ fixed. The accumulated drift is relatively slow compared to the dynamics of an aerial robot. We therefore optimize the graph at a lower rate of \SI{10}{\hertz}, allowing us to infer the current drift over a larger sliding window of \SI{20}{s} and increase the accuracy of the estimate while still keeping the computational cost low.   

\vspace{-5pt}
\subsection{Consensus finding}
\begin{figure} %
    \centering
    \includegraphics[width=0.8\linewidth]{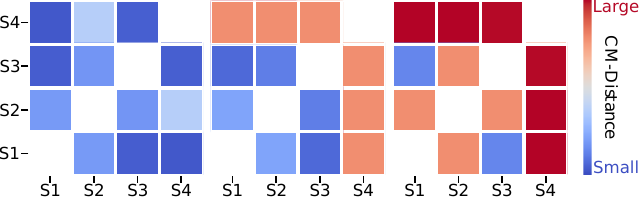}
    \caption{Illustration of the consistency matrix for the case of four sensors. From left to right: no sensor failure, failure of S4, and failure of S2 and S4. }
    \vspace{-10pt}
    \label{fig:consensus}
\end{figure}

\vspace{-5pt}
We use the above-mentioned metrics for consensus finding when employing multiple parallel estimators. The metric should be a distance that represents the consistency between the measurements of all different sensor combinations. Based on these consistency values, we determine the most consistent sensor pair and use measurements of one of the two sensors as constraints in our local estimate. Additionally, it allows us to determine if a sensor provides corrupted measurements or has failed completely. \Cref{fig:consensus} visualizes the consistency values in matrix form for four sensors and an increasing number of sensor failures from left to right.
In case of sensor inconsistencies, all off-diagonal values corresponding to the sensor's row or column increase rapidly, while the other values remain low, resulting in a cross-pattern. As long as more than two sensors are available, we can exclude pose estimates coming from a corrupted sensor by searching the consistency matrix for values exceeding a predefined threshold within this pattern. For the case of two inconsistent sensors, however, the distance metric does not provide any information on which sensor is faulty and which is not. To determine a valid sensor in such a degenerate case, we additionally compute the consistency distance between each sensor and the local estimate. A faulty sensor will corrupt the local estimate when fused, resulting in both sensors being inconsistent compared to the local estimate. Conversely, when a healthy sensor is fused, it will remain consistent with the local estimate, while the other sensor's inconsistency will increase.
\vspace{-5pt}
\section{Experiments}
\vspace{-5pt}
\label{sec:experiments}

We start with indoor experiments, that allow us to answer which of the proposed metrics is best suited to detect sensor divergence and evaluate which failure modes can be detected and mitigated by our sensor fusion framework. Then we perform experiments using data from real-world outdoor deployments that showcase the performance of our complete pipeline in realistic scenarios.

\vspace{-8pt}
\subsection{Indoor Dataset}
\vspace{-5pt}
We imitate visual inspection tasks near structures using an omnidirectional aerial robot indoors which provides us with ground truth data from a Vicon system for comparison purposes. The robot is equipped with an Adis 16448B IMU, an Intel Realsense T265, a Livox Mid-360 Lidar, and a GRZ101 \SI{360}{\degree} Mini Prism tracked using a Leica MS60 total station. Fast-LIO \cite{xu2021fastlio} is used to provide odometry measurements.
All sensors provide either odometry or position measurements at different rates and with different noise and accuracy characteristics.

\vspace{-5pt}
\subsection{Metric Evaluation}
\vspace{-3pt}
\begin{wrapfigure}{r}{0.5\textwidth}
    \vspace{-35pt}  
    \centering
    \includegraphics[trim=0 15 0 0, clip, width=\linewidth]{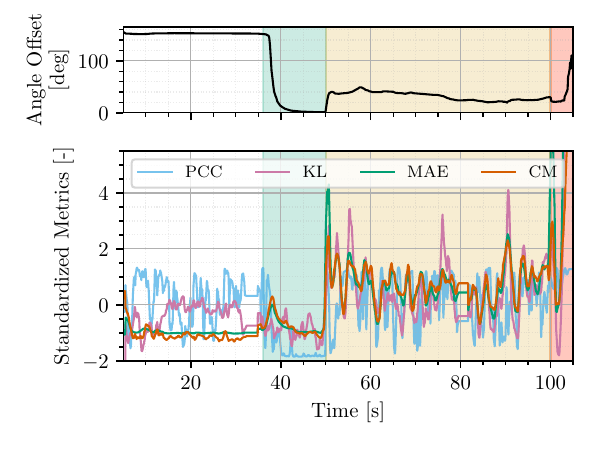}
    \vspace{-20pt}
    \caption{Angular difference between the true and estimated orientation (top) and the different standardized metrics shown (bottom). The green region has all states observable, yellow has GNSS-like noise artificially applied, and red has the sensor data diverging.}
    \vspace{-20pt}
    \label{fig:metrics}
\end{wrapfigure}

To evaluate the suitability of the different metrics we artificially corrupt the sensor measurements simulating the following failure modes: i) incorrect reference frame alignment, ii) high measurement noise, and iii) diverging estimates. The data is corrupted in the following manner.
The orientation estimate is initialized with an offset of \SI{30}{\degree}, \SI{60}{\degree}, and \SI{120}{\degree} in roll, pitch, and yaw respectively. To simulate realistic GNSS measurements, the total station measurements are truncated after \SI{50}{\s} with non-Gaussian noise composed of white and brown noise with a similar magnitude to GNSS. After \SI{100}{s}, artificial measurement drift is introduced by integrating an exponentially increasing velocity along the platform's y-axis and adding it to the position measurement. 

Every metric uses velocity measurements obtained within the last \SI{1}{\s} and centered and scaled to the same range. \Cref{fig:metrics} shows the angular difference between the true and estimated Leica reference frame as well as the performance of the different metrics during the experiment.
In the first \SI{35}{\s} (white), the platform takes off and only moves along the $z$ axis. While the estimated orientation converges in roll and pitch, the yaw offset remains unobservable, resulting in a large angular offset. Only when the platform moves forward (green), at \SI{35}{\s}, are the remaining inconsistencies in the transformation observable.

\begin{wrapfigure}{r}{0.5\textwidth}
    \vspace{-20pt}  
    \centering
    \includegraphics[trim=0 30 0 0, clip,width=\linewidth]{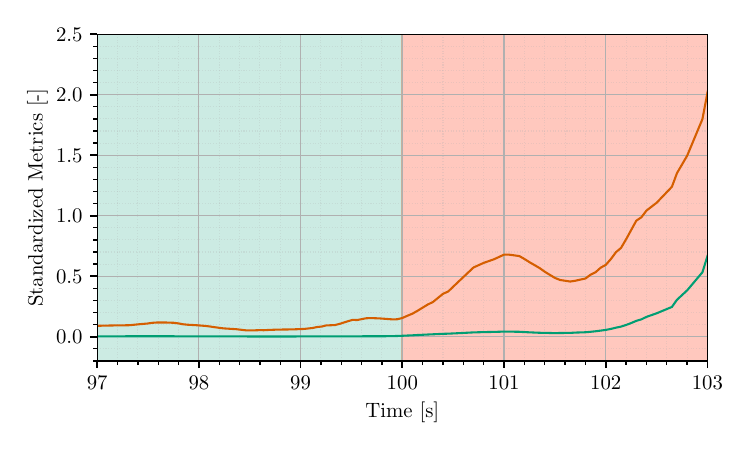}
    \caption{A zoomed-in example transitioning from accurate (green background) to simulated diverging (red background) measurement. CM (orange) responds faster than MAE (green).}
    \vspace{-20pt}
    \label{fig:slow_mae}
\end{wrapfigure}

All metrics have low values (consistent) in the green section of the graph when high-accuracy data is available, but their response to the degenerate scenarios differs significantly.
The PCC, for example, is equally large in the presence of mild noise (yellow) and complete divergence (red). As the measurement noise is larger than the linear relationship between the two velocities the metric is unsuitable.

The KL Divergence also does poorly in capturing the agreement between two sensors, such as in the green region. This is because the metric compares two \textit{distributions}, which we estimate using \ac{KDE}, and is therefore highly sensitive to \ac{KDE} hyperparameters, which must be fit per sensor. This is impractical in our scenario. Finally, both MAE and Cram\'er distance reflect the usability of the measurement. However, as shown in \Cref{fig:slow_mae}, the \ac{CM}-Distance responds much faster when the estimate diverges. Since MAE acts as a low-pass filter, it introduces delays, while \ac{CM} can handle outliers instantaneously.

As a result of this analysis, we have selected the \ac{CM}-Distance as the metric we will consider for the remainder of our work.

\vspace{-5pt}
\subsection{Consensus-based Sensor Selection}
\begin{figure}[tb]
    \centering
    \includegraphics[width=0.9\linewidth]{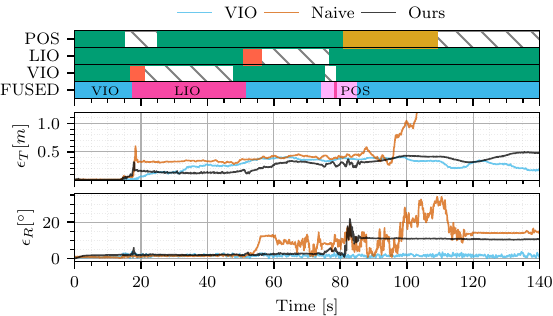}
    \caption{Evaluation of various fusion approaches on the indoor dataset with corrupted sensor data. Top shows which corruption is applied when: green is uncorrupted, white is drop-out, yellow is non-Gaussian noise, and red is divergence. The results of fusing only \textit{uncorrupted} VIO pose estimates is shown in blue, and the naive approach of fusing all \textit{corrupted} sensors is shown in orange. Our consensus-based fusion approach is shown in black, with the ``FUSED'' line showing which sensor is actively fused at any time.}
    \label{fig:abs_errs}
    \vspace{-15pt}
\end{figure}

\vspace{-3pt}
In the following, we evaluate the performance of our CM-distance based selective sensor fusion framework in comparison to naively fusing all sensor data. 
\cref{fig:abs_errs} shows how the sensor data is corrupted (top) and how the various fusion methods handle the corrupted data in position error (middle) and rotational error (bottom) compared to the Vicon ground truth.
In the top section, green shows uncorrupted regions of the Leica (POS) position, LiDAR-inertial (LIO), and visual-inertial (VIO) pose measurements; white sections indicate complete drop-outs; yellow is corrupted with non-Gaussian noise; red is diverging.

The middle and bottom graphs show the results of three different fusion strategies. VIO, in blue, shows the result of fusing \textit{only} the \textit{uncorrupted} VIO pose estimate, acting as a best-possible realistic estimate with high quality. In orange, we naively fuse all \textit{corrupted} sensors, and in black is our consensus-based fusion approach on the \text{corrupted} data. The ``FUSED'' line shows which sensors are currently actively fused in our approach.

We can clearly see that our consensus-based approach shows far lower errors than fusing all sensors naively. Also note that in the naive case, we use the ground truth transform between the Leica and local frame, which we cannot estimate accurately. Even with this privileged information, the estimate drifts extensively after 90 seconds when non-Gaussian noise is fused and only stabilizes when the position measurements drop out.

For the implementation of the consensus-based fusion, we set a threshold of \SI[per-mode=symbol]{0.1}{\metre\per\s} for the CM-Distance metric. This value was chosen as a trade-off between not too frequent sensor switching (larger value) and detecting divergences and degenerate measurements rapidly (small value), but note that the performance of the overall algorithm is quite sensitive to this threshold. 

While frequent switching usually does not cause jumps or inconsistencies in the output of the local graph estimate, it is associated with a certain risk that a degenerate sensor is chosen. 
This can occur when only two ``valid" sensors are available, and the used sensor is chosen based on the local estimate. In such a case, for a too strict threshold, the consensus algorithm keeps switching between a valid and a corrupted sensor, resulting in a noisy local output, making it impossible to converge to the correct sensor based solely on the local estimate. 

In general, our approach is quicker to recover from drifts (such as at 18 seconds when VIO diverges), and while both initially struggle with non-Gaussian noise on the Leica (82 seconds), our approach is able to recover and keep a bounded error while the naive approach drifts away completely.

\vspace{-10pt}
\subsection{Outdoor datasets}
\vspace{-5pt}
We evaluate our overall fusion approach in two outdoor scenarios: transitioning between indoors and outdoors by flying through a window in a partially destroyed building and flying under a bridge while performing \ac{NDT}.
In both datasets, the robot uses visual-inertial odometry and a source of global position: in the case of the indoor-outdoor dataset, RTK GNSS, and in the bridge case, a Leica total station.

\begin{figure}[tb]
    \vspace{-10pt}
    \centering
    \includegraphics[width=\linewidth]{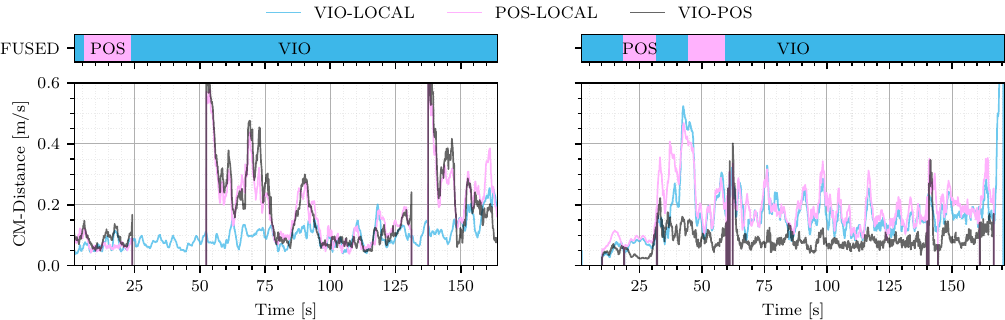}
    \caption{The CM-metric for the indoor-outdoor (left) and bridge-inspection dataset (right). Measurement dropouts are marked by a negative value CM-metric. The top bars show at which time what sensor is actively being fused. }
    \label{fig:cm_qual}
    \vspace{-15pt}
\end{figure}

\cref{fig:cm_qual} shows the output of the \ac{CM} metric and which sensors were being actively fused by our framework at which time. 
The left graph shows the indoor-outdoor scenario, which contains several cases of losing the RTK GNSS, which are immediately caught by the CM metric. Both the POS-LOCAL metric, measuring how consistent the GNSS and the IMU are, and the VIO-POS metric, measuring consistency between VIO and GNSS become much larger when this happens.
This allows our fusion approach to automatically switch to fusing mainly VIO when these outages occur.
However, when the robot leaves the building at \SI{50}{s}, the estimator continues to use VIO. The RTK is still searching for a fix until approximately \SI{130}{s} when the estimator briefly switches to using the GNSS, but then the metric between the local and GNSS estimate becomes too high briefly, and it switches back to VIO for the rest of the dataset.
This is because we used the same switching threshold as in the quantitative experiments, \SI[per-mode=symbol]{0.1}{\metre\per\s}, making it quite noise-sensitive.

The bridge inspection dataset (\cref{fig:cm_qual} right) features an omnidirectional aerial vehicle flying under a bridge.
The dataset features an unsuccessful trial where the on-board state estimate diverges at the end. For this dataset, the \textit{local} estimate is generally poor due to insufficient damping on the \acp{IMU}, resulting in large \ac{IMU} noise. For this reason, we increased the inconsistency threshold to \SI[per-mode=symbol]{0.3}{\metre\per\s} for this dataset. Despite this the metric was still able to accurately capture when the Leica tracking was lost due to jerky flight motions: at \SI{60}{s} and \SI{140}{s}.
At the end of the dataset, the VIO estimate diverges due to extremely high IMU noise. Our metric also captures this: VIO-LOCAL increases rapidly, showing that that sensor combination can no longer be used.

Our qualitative results show that the CM-Distance metric is informative about the state of the sensors in real-world degenerate sensor scenarios.

\vspace{-10pt}
\section{Conclusion}
\vspace{-5pt}
In this work we evaluated several metrics for state estimation consistency assessment and selected the \acf{CM} Distance as the most suitable. Based on this we developed a consensus-based sensor fusion framework and evaluated it on artificially corrupted and challenging real-world datasets. The results demonstrate that we can accurately detect inconsistent sensors and that selectively fusing sensors outperforms naively fusing all available data.
\vspace{-5pt}

\bibliographystyle{ieeetr}
\bibliography{references}
\end{document}